\documentclass[english]{article}
\usepackage[T1]{fontenc}
\usepackage[latin9]{inputenc}
\usepackage{babel}
\usepackage{array}
\usepackage{multirow}
\usepackage{amsmath}
\usepackage[unicode=true]
 {hyperref}
\usepackage{breakurl}

\makeatletter

\providecommand{\tabularnewline}{\\}

\usepackage{times}
\DeclareMathOperator*{\T}{\scriptscriptstyle \top}
\title{Dirichlet-vMF Mixture Model}
\author{Shaohua Li\\shaohua@gmail.com\\National University of Singapore}

\makeatother

\begin{document}
\maketitle
\begin{abstract}
This document is about the multi-document Von-Mises-Fisher mixture
model with a Dirichlet prior, referred to as VMFMix. VMFMix is analogous
to Latent Dirichlet Allocation (LDA) in that they can capture the
co-occurrence patterns acorss multiple documents. The difference is
that in VMFMix, the topic-word distribution is defined on a continuous
n-dimensional hypersphere. Hence VMFMix is used to derive topic embeddings,
i.e., representative vectors, from multiple sets of embedding vectors.
An efficient Variational Expectation-Maximization inference algorithm
is derived. The performance of VMFMix on two document classification
tasks is reported, with some preliminary analysis. 
\end{abstract}
We present a simplification of the Bayesian vMF mixture model proposed
in \cite{vmfmix}\footnote{This model reappears in \cite{vmftopic} under the name ``mix-vMF
topic model''. But \cite{vmftopic} only offers a sampling-based
inference scheme, which is usually less accurate than the EM algorithm
presented in this document.}. For computational efficiency, the priors on the vMF mean $\{\boldsymbol{\mu}_{k}\}$
and on the vMF concentration $\{\kappa_{k}\}$ are removed. This model
is referred to as \textbf{VMFMix}. 

A Python implementation of VMFMix is available at \href{https://github.com/askerlee/vmfmix}{https://github.com/askerlee/vmfmix}.

\section{Model Specification}

The generative process is as follows:
\begin{enumerate}
\item $\boldsymbol{\theta}_{i}\sim\textrm{Dir}(\alpha)$;
\item $z_{ij}\sim\textrm{Cat}(\boldsymbol{\theta}_{i})$;
\item $\boldsymbol{x}_{ij}\sim\textrm{vMF}(\boldsymbol{\mu}_{z_{ij}},\kappa_{z_{ij}})$.
\end{enumerate}
Here $\alpha$ is a hyperparameter, $\{\boldsymbol{\mu}_{k},\kappa_{k}\}$
are parameters of mixture components to be learned.

\section{Model Likelihood and Inference}

Given parameters $\{\boldsymbol{\mu}_{k},\kappa_{k}\}$, the complete-data
likelihood of a dataset $\{\boldsymbol{X},\boldsymbol{Z},\boldsymbol{\Theta}\}=\{\boldsymbol{x}_{ij},z_{ij},\boldsymbol{\theta}_{i}\}$
is:

\begin{equation}
p(\boldsymbol{X},\boldsymbol{Z},\boldsymbol{\Theta}|\alpha,\{\boldsymbol{\mu}_{k},\kappa_{k}\})=\prod_{i}\textrm{Dir}(\boldsymbol{\theta}_{i}|\alpha)\prod_{j}\theta_{i,z_{ij}}\textrm{vMF}(\boldsymbol{x}_{ij}|\boldsymbol{\mu}_{z_{ij}},\kappa_{z_{ij}}).\label{eq:complete-like}
\end{equation}

The incomplete-data likelihood of $\{\boldsymbol{X},\boldsymbol{\Theta}\}=\{\boldsymbol{x}_{ij},\boldsymbol{\theta}_{i}\}$
is obtained by integrating out the latent variables $\boldsymbol{Z},\boldsymbol{\Theta}$:

\begin{equation}
p(\boldsymbol{X}|\alpha,\{\boldsymbol{\mu}_{k},\kappa_{k}\})=\int d\boldsymbol{\Theta}\cdot\prod_{i}\textrm{Dir}(\boldsymbol{\theta}_{i}|\alpha)\prod_{j}\sum_{k}\theta_{ik}\textrm{vMF}(\boldsymbol{x}_{ij}|\boldsymbol{\mu}_{k},\kappa_{k}).\label{eq:incomp-like}
\end{equation}

(\ref{eq:incomp-like}) is apparently intractable, and instead we
seek its variational lower bound:

\begin{align}
\log p(\boldsymbol{X}|\alpha,\{\boldsymbol{\mu}_{k},\kappa_{k}\}) & \ge E_{q(\boldsymbol{Z},\boldsymbol{\Theta})}[\log p(\boldsymbol{X},\boldsymbol{Z},\boldsymbol{\Theta}|\alpha,\{\boldsymbol{\mu}_{k},\kappa_{k}\})-\log q(\boldsymbol{Z},\boldsymbol{\Theta})].\nonumber \\
 & =\mathcal{L}(q,\{\boldsymbol{\mu}_{k},\kappa_{k}\})\label{eq:varlb}
\end{align}

It is natural to use the following variational distribution to approximate
the posterior distribution of $\boldsymbol{Z},\boldsymbol{\Theta}$:

\begin{equation}
q(\boldsymbol{Z},\boldsymbol{\Theta})=\prod_{i}\Bigl\{\textrm{Dir}(\boldsymbol{\theta}_{i}|\boldsymbol{\phi}_{i})\prod_{j}\textrm{Cat}(z_{ij}|\boldsymbol{\pi}_{ij})\Bigr\}.\label{eq:q}
\end{equation}

Then the variational lower bound is
\begin{align}
 & \mathcal{L}(q,\{\boldsymbol{\mu}_{k},\kappa_{k}\})\nonumber \\
= & C_{0}+\mathcal{H}(q)+E_{q(\boldsymbol{Z},\boldsymbol{\Theta})}\Bigl[(\alpha-1)\sum_{i,k}\log\theta_{ik}\nonumber \\
 & +\sum_{i,j,k}\delta(z_{ij}=k)(\log\theta_{ik}+\log c_{d}(\kappa_{k})+\kappa_{k}\boldsymbol{\mu}_{k}^{\T}\boldsymbol{x}_{ij})\Bigr]\nonumber \\
= & C_{0}+\mathcal{H}(q)+\sum_{i,k}(\alpha-1+n_{i\cdot k})\Bigl(\psi(\phi_{ik})-\psi(\phi_{i0})\Bigr)\nonumber \\
 & +\sum_{k}\Bigl(n_{\cdot\cdot k}\cdot\log c_{d}(\kappa_{k})+\kappa_{k}\boldsymbol{\mu}_{k}^{\T}\boldsymbol{r}_{k}\Bigr),\label{eq:varlb2}
\end{align}

where
\begin{align}
n_{i\cdot k} & =\sum_{j}\pi_{ijk},\quad n_{\cdot\cdot k}=\sum_{i,j}\pi_{ijk},\label{eq:n}\\
\boldsymbol{r}_{k} & =\sum_{i,j}\pi_{ijk}\cdot\boldsymbol{x}_{ij},\label{eq:rk}
\end{align}
and $\mathcal{H}(q)$ is the entropy of $q(\boldsymbol{Z},\boldsymbol{\Theta})$:
\begin{align}
\mathcal{H}(q)= & -E_{q}[\log q(\boldsymbol{Z},\boldsymbol{\Theta})]\nonumber \\
= & \sum_{i}E_{q}\Bigl[\sum_{k}\log\Gamma(\phi_{ik})-\log\Gamma(\phi_{i0})-\sum_{k}(\phi_{ik}-1)\log\theta_{ik}\nonumber \\
 & -\sum_{j,k}\delta(z_{ij}=k)\log\pi_{ijk}\Bigr]\nonumber \\
= & \sum_{i}\Bigl(\sum_{k}\log\Gamma(\phi_{ik})-\log\Gamma(\phi_{i0})-\sum_{k}(\phi_{ik}-1)\psi(\phi_{ik})\Bigr)\nonumber \\
 & +(\phi_{i0}-K)\psi(\phi_{i0})-\sum_{j,k}\pi_{ijk}\log\pi_{ijk}.\label{eq:Hq}
\end{align}

By taking the partial derivative of (\ref{eq:varlb2}) w.r.t. $\pi_{ijk},\phi_{ik},\boldsymbol{\mu}_{k},\kappa_{k},$
respectively, we can obtain the following variational EM update equations
\cite{vmfclust,vmfmix,vmftopic}.

\subsection{E-Step}

\begin{align}
\pi_{ijk} & \sim e^{\psi(\phi_{ik})}\cdot\textrm{vMF}(\boldsymbol{x}_{ij}|\boldsymbol{\mu}_{k},\kappa_{k}),\nonumber \\
\phi_{ik} & =n_{i\cdot k}+\alpha.\label{eq:e-step}
\end{align}

\subsection{M-Step}

\begin{align}
\boldsymbol{\mu}_{k} & =\frac{\boldsymbol{r}_{k}}{\left\Vert \boldsymbol{r}_{k}\right\Vert },\nonumber \\
\bar{r}_{k} & =\frac{\left\Vert \boldsymbol{r}_{k}\right\Vert }{n_{..k}},\nonumber \\
\kappa_{k} & \approx\frac{\bar{r}_{k}D-\bar{r}_{k}^{3}}{1-\bar{r}_{k}^{2}}.\label{eq:m-step}
\end{align}

The update equation of $\kappa_{k}$ adopts the approximation proposed
in \cite{vmfclust}.

\section{Evaluation}

The performance of this model was evaluated on two text classification
tasks that are on 20 Newsgroups (\textbf{20News}) and \textbf{Reuters},
respectively. The experimental setup for the compared methods were
identical to that in \cite{topicvec}. Similar to TopicVec, VMFMix
learns an individual set of $K$ topic embeddings from each category
of documents, and all these sets are combined to form a bigger set
of topic embeddings for the whole corpus. This set of topic embeddings
are used to derive the topic proportions of each document, which are
taken as features for the SVM classifier. The $K$ for 20News and
Reuters are chosen as 15 and 12, respectively, which are identical
to TopicVec. 

The macro-averaged precision, recall and F1 scores of all methods
are presented in Table 1.

\begin{table}
\centering{}\renewcommand{\thempfootnote}{\arabic{mpfootnote}}%
\noindent\begin{minipage}[t]{1\columnwidth}%
\begin{center}
\begin{tabular}{|c|c|c|c|c|c|c|}
\hline 
\multirow{2}{*}{} & \multicolumn{3}{c|}{{\footnotesize{}20News}} & \multicolumn{3}{c|}{{\footnotesize{}Reuters}}\tabularnewline
\cline{2-7} 
 & {\footnotesize{}Prec} & {\footnotesize{}Rec} & {\footnotesize{}F1} & {\footnotesize{}Prec} & {\footnotesize{}Rec} & {\footnotesize{}F1}\tabularnewline
\hline 
\hline 
{\footnotesize{}BOW} & {\footnotesize{}69.1} & {\footnotesize{}68.5} & {\footnotesize{}68.6} & {\footnotesize{}92.5} & {\footnotesize{}90.3} & {\footnotesize{}91.1}\tabularnewline
\hline 
{\footnotesize{}LDA} & {\footnotesize{}61.9} & {\footnotesize{}61.4} & {\footnotesize{}60.3} & {\footnotesize{}76.1} & {\footnotesize{}74.3} & {\footnotesize{}74.8}\tabularnewline
\hline 
{\footnotesize{}sLDA} & {\footnotesize{}61.4} & {\footnotesize{}60.9} & {\footnotesize{}60.9} & {\footnotesize{}88.3} & {\footnotesize{}83.3} & {\footnotesize{}85.1}\tabularnewline
\hline 
{\footnotesize{}LFTM} & {\footnotesize{}63.5} & {\footnotesize{}64.8} & {\footnotesize{}63.7} & {\footnotesize{}84.6} & {\footnotesize{}86.3} & {\footnotesize{}84.9}\tabularnewline
\hline 
{\footnotesize{}MeanWV} & {\footnotesize{}70.4} & {\footnotesize{}70.3} & {\footnotesize{}70.1} & {\footnotesize{}92.0} & {\footnotesize{}89.6} & {\footnotesize{}90.5}\tabularnewline
\hline 
{\footnotesize{}Doc2Vec} & {\footnotesize{}56.3} & {\footnotesize{}56.6} & {\footnotesize{}55.4} & {\footnotesize{}84.4} & {\footnotesize{}50.0} & {\footnotesize{}58.5}\tabularnewline
\hline 
{\footnotesize{}TWE} & {\footnotesize{}69.5} & {\footnotesize{}69.3} & {\footnotesize{}68.8} & {\footnotesize{}91.0} & {\footnotesize{}89.1} & {\footnotesize{}89.9}\tabularnewline
\hline 
{\footnotesize{}TopicVec} & \textbf{\footnotesize{}71.3} & \textbf{\footnotesize{}71.3} & \textbf{\footnotesize{}71.2} & \textbf{\footnotesize{}92.5} & \textbf{\footnotesize{}92.1} & \textbf{\footnotesize{}92.2}\tabularnewline
\hline 
{\footnotesize{}VMFMix} & {\footnotesize{}63.8} & {\footnotesize{}63.9} & {\footnotesize{}63.7} & {\footnotesize{}87.9} & {\footnotesize{}88.7} & {\footnotesize{}88.0}\tabularnewline
\hline 
\end{tabular}
\par\end{center}%
\end{minipage}\caption{Performance on multi-class text classification. Best score is in boldface.}
\end{table}

We can see from Table 1 that, VMFMix achieves better performance than
Doc2Vec, LDA, sLDA and LFTM. However, its performance is still inferior
to BOW, Mean word embeddings (MeanWV), TWE and TopicVec. The reason
might be that by limiting the embeddings in the unit hypersphere (effectively
normalizing them as unit vectors), certain representational flexibility
is lost.

An empirical observation we have is that, VMFMix approaches convergence
very quickly. The variational lower bound increases only slightly
after 10\textasciitilde{}20 iterations. By manually checking the intermediate
parameter values, we see that after so many iterations, the parameters
change very little too. It suggests that VMFMix might easily get stuck
in local optima.

Nonetheless, VMFMix might still be relevant when the considered embedding
vectors are infinite and continuously distributed in the embedding
space, as opposed to the finite vocabulary of word embeddings\footnote{Each set of word embeddings can be viewed as a finite and \emph{discrete}
sample from a \emph{continuous} embedding space.}. Such scenarios include the neural encodings of images from a convolutional
neural network (CNN).

\bibliographystyle{plain}
\bibliography{vMF_mixture}

\begin{thebibliography}{1}

\bibitem{vmfclust}
Arindam Banerjee, Inderjit~S Dhillon, Joydeep Ghosh, and Suvrit Sra.
\newblock Clustering on the unit hypersphere using von mises-fisher
  distributions.
\newblock {\em Journal of Machine Learning Research}, 6(Sep):1345--1382, 2005.

\bibitem{vmfmix}
Siddharth Gopal and Yiming Yang.
\newblock Von mises-fisher clustering models.
\newblock In {\em ICML}, pages 154--162, 2014.

\bibitem{topicvec}
Shaohua Li, Tat{-}Seng Chua, Jun Zhu, and Chunyan Miao.
\newblock Generative topic embedding: a continuous representation of documents.
\newblock In {\em Proceedings of the 54th Annual Meeting of the Association for
  Computational Linguistics, {ACL} 2016, August 7-12, 2016, Berlin, Germany,
  Volume 1: Long Papers}, 2016.

\bibitem{vmftopic}
Ximing Li, Jinjin Chi, Changchun Li, Jihong OuYang, and Bo~Fu.
\newblock Integrating topic modeling with word embeddings by mixtures of vmfs.
\newblock In {\em COLING}, 2016.

\end{thebibliography}

\end{document}